\documentclass[sw,crcready]{iosart2x}

\usepackage{xspace}
\usepackage{xcolor,listings}
\usepackage{hyperref}
\definecolor{darkblue}{rgb}{0, 0, 0.5}
\hypersetup{colorlinks=true, citecolor=darkblue, linkcolor=darkblue, urlcolor=darkblue}

\newcommand{\deeponto}{\textsf{DeepOnto}\xspace}
\usepackage{multirow}
\usepackage{tabularx}
\usepackage{textcomp}
\usepackage{listings}
\usepackage{booktabs}
\newcolumntype{Y}{>{\centering\arraybackslash}X}

\usepackage{tcolorbox}
\newtcbox{\inlinecode}{on line, boxrule=0pt, boxsep=0pt, top=2pt, left=2pt, bottom=2pt, right=2pt, colback=gray!15, colframe=white, fontupper={\ttfamily \footnotesize}}

\usepackage{titlesec}

\newcommand{\nsp}{\negthickspace}

\titleformat{\subsubsection}[runin]{\normalfont\normalsize\bfseries}{\thesubsubsection}{1em}{}[]

\titlespacing{\subsubsection}{0pt}{1.0ex plus 1.0ex minus .2ex}{1.5ex plus .2ex}

\pubyear{0000}
\volume{0}
\firstpage{1}
\lastpage{1}

\begin{document}

\begin{frontmatter}
 
\title{DeepOnto: A Python Package for Ontology Engineering with Deep Learning}
\runtitle{DeepOnto}


\begin{aug}
\author[A]{\fnms{Yuan} \snm{He}\ead[label=e1]{yuan.he@cs.ox.ac.uk}}
\author[B,A]{\fnms{Jiaoyan} \snm{Chen}\ead[label=e2]{jiaoyan.chen@manchester.ac.uk}}
\author[A]{\fnms{Hang} \snm{Dong}\ead[label=e3]{h.dong2@exeter.ac.uk}}
\author[A]{\fnms{Ian} \snm{Horrocks}\ead[label=e7]{ian.horrocks@cs.ox.ac.uk}}
\author[C]{\fnms{Carlo} \snm{Allocca}\ead[label=e4]{c.allocca@samsung.com}}
\author[C]{\fnms{Taehun} \snm{Kim}\ead[label=e5]{taehun11.kim@samsung.com}}
\author[C]{\fnms{Brahmananda} \snm{Sapkota}\ead[label=e6]{b.sapkota@samsung.com}}
\address[A]{Department of Computer Science, \orgname{University of Oxford}, \cny{UK}\printead[presep={\\}]{e1,e3,e7}}
\address[B]{Department of Computer Science, \orgname{The University of Manchester},\cny{UK}\printead[presep={\\}]{e2}}
\address[C]{\orgname{Samsung Research}, \cny{UK}\printead[presep={\\}]{e4,e5,e6}}
\end{aug}

\begin{abstract}

Integrating deep learning techniques, particularly language models (LMs), with knowledge representation techniques like ontologies has raised widespread attention, urging the need of a platform that supports both paradigms. Although packages such as OWL API and Jena offer robust support for basic ontology processing features, they lack the capability to transform various types of information within ontologies into formats suitable for downstream deep learning-based applications. Moreover, widely-used ontology APIs are primarily Java-based while deep learning frameworks like PyTorch and Tensorflow are mainly for Python programming. To address the needs, we present \deeponto, a Python package designed for ontology engineering with deep learning. The package encompasses a core ontology processing module founded on the widely-recognised and reliable OWL API, encapsulating its fundamental features in a more ``Pythonic'' manner and extending its capabilities to incorporate other essential components including reasoning, verbalisation, normalisation, taxonomy, projection, and more. Building on this module, \deeponto offers a suite of tools, resources, and algorithms that support various ontology engineering tasks, such as ontology alignment and completion, by harnessing deep learning methods, primarily pre-trained LMs. In this paper, we also demonstrate the practical utility of \deeponto through two use-cases: the Digital Health Coaching in Samsung Research UK and the Bio-ML track of the Ontology Alignment Evaluation Initiative (OAEI).
\end{abstract}


\end{frontmatter}

\vspace{-.8cm}
{\small
\noindent\textbf{Repository}: \url{https://github.com/KRR-Oxford/DeepOnto}

\noindent\textbf{Documentation}: \url{https://krr-oxford.github.io/DeepOnto/}

\noindent\textbf{License}: \href{https://github.com/KRR-Oxford/DeepOnto/blob/main/LICENSE}{Apache License, Version 2.0}
}


\section{Introduction} \label{section:intro}

An ontology is a formal, explicit specification of knowledge within the scope of a domain. It provides a vocabulary of concepts and properties that enables a shared understanding of semantics among humans and machines, with wide applications in many domains such as bioinformatics, information systems, and the Semantic Web.
Ontology engineering, a sub-field of knowledge engineering, underpins the various stages of ontology development, encompassing ontology design, construction, curation, evaluation, and maintenance, among others \cite{staab2010handbook}. Concrete tasks of ontology engineering include: \textit{(i)} 
defining the entities and constructing the logical axioms that make up an ontology, \textit{(ii)} validating and ensuring quality (e.g., completeness and correctness) of an ontology, \textit{(iii)} inserting new knowledge into an ontology, \textit{(iv)} integrating domain ontologies that come from heterogeneous sources, and so on. These tasks can collectively enhance an ontology's practical utility, making it applicable to different real-world scenarios.

Deep learning approaches have gained significant popularity across various research and engineering domains. These techniques have shown notable advantages over conventional ontology engineering tools. For example, LogMap \cite{jimenez2011logmap}, a long-standing state-of-the-art ontology alignment system, relies on lexical similarity and lacks the ability to capture textual contexts. In contrast, BERTMap \cite{he2022bertmap}, a language model (LM)-based system which leverages the attention mechanism of the transformer architecture for contextual text embeddings \cite{vaswani2017attention}, can be more robust to linguistic variations such as synonyms and polysemies. Another example concerns ontology completion. Traditional systems, leveraging formal logics and/or heuristic rules, are capable of inferring entailed knowledge (e.g., HermiT for ontology reasoning \cite{glimm2014hermit}). But the requirement of manual design and curation for these logics and rules frequently results in their under-specification. This constraint significantly curtails their capability to deduce absent but plausible knowledge. As opposed to these rule-based solutions, deep learning-based techniques can automatically learn patterns from the existing knowledge, the metadata and different kinds of other information, and make predictions accordingly~\cite{li2019ontology,chen2023contextual,chen2021owl2vec}. Nevertheless, the benefit is not single sided. The formal and structural semantics embedded in ontologies can augment deep learning models, enhancing not only their training efficiency but also their interpretability. For instance, leveraging structural contexts via graph-based attention mechanisms has yielded substantial advancements in predictive modelling \cite{choi2017gram}. Furthermore, the issue of mitigating hallucinated responses produced by recent large language models (LLMs) can be addressed through the incorporation of attributions \cite{rashkin2023measuring,bohnet2022attributed}, potentially from symbolic knowledge. In this context, ontologies can play a crucial role, acting as robust and trustworthy reference points to validate and support the generated answers.

However, there lacks a systematic support for integrating ontology engineering with deep learning, posing challenges for both developers and users. While there are packages such as OWL API \cite{owlapi} and Jena \cite{carroll2004jena} that effectively support basic ontology processing features, particularly those associated with OWL (Web Ontology Language) \cite{motik2009owl}--a prominent ontology language grounded in Description Logic--to the best of our knowledge, no existing packages are designed to transform various types of information within ontologies into formats to facilitate a broad spectrum of deep learning-based ontology engineering solutions. This is further complicated by the fact that leading deep learning frameworks like Tensorflow \cite{abadi2016tensorflow} and PyTorch \cite{paszke2019pytorch} primarily offer Python-based APIs, whereas the majority of ontology APIs are Java-based. Although the mOWL toolkit \cite{zhapa2023mowl} resolves this gap by employing JPype\footnote{\url{https://jpype.readthedocs.io/en/latest/install.html}} to bridge Python and Java, it mainly focuses on ontology embedding models and their applications to the biomedical domain, thereby neglecting fundamental ontology processing features and a variety of essential tasks pivotal for ontology construction and curation. In order to provide a comprehensive, general, and versatile package for supporting deep learning-based ontology engineering, we develop \deeponto. 

\begin{figure}[t!]
\begin{center}
    \includegraphics[width=0.99\textwidth]{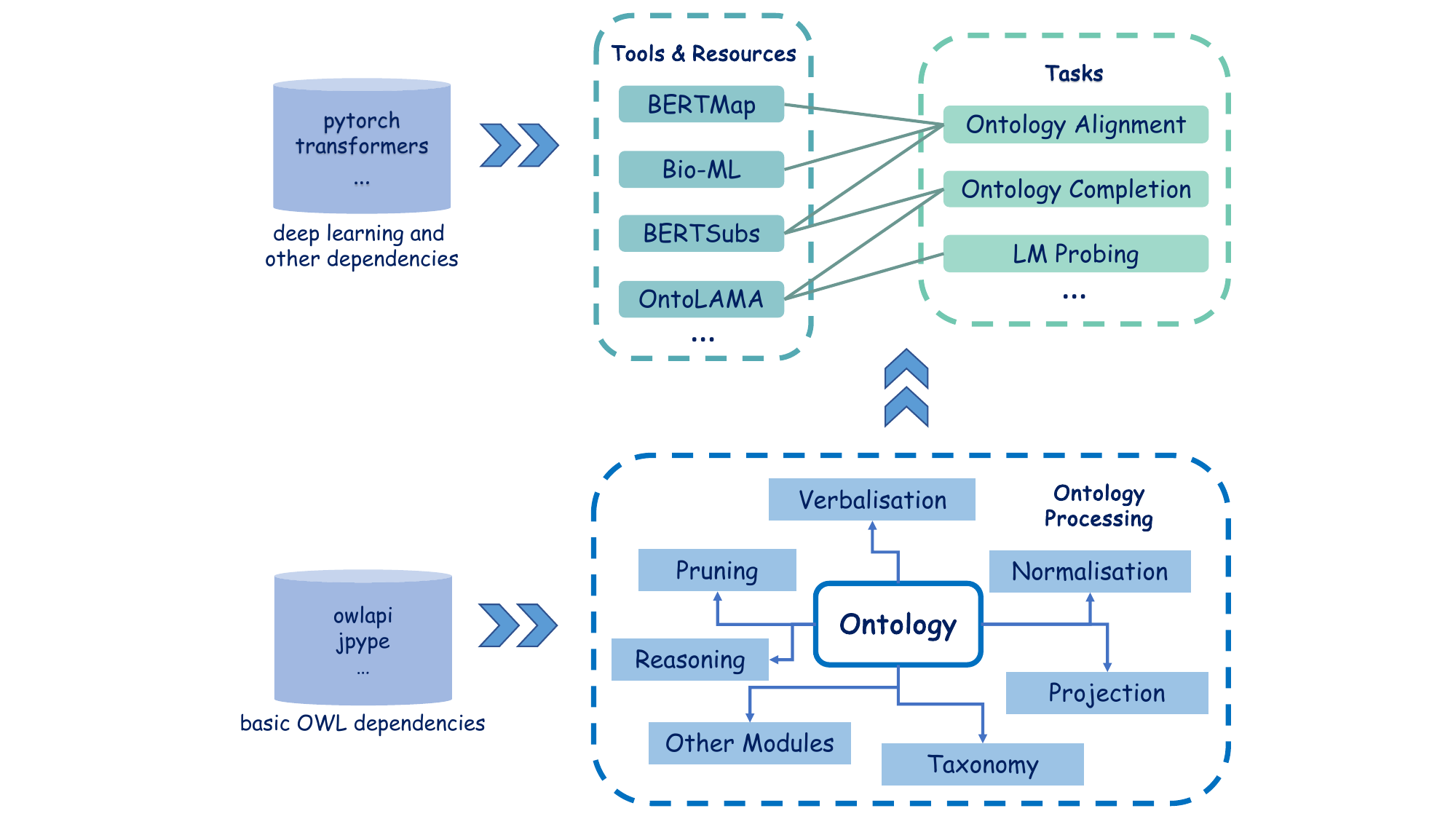}
\caption{Illustration of \deeponto's architecture, with the lower half depicting the core ontology processing module, and the upper half presenting various tools and resources for diverse ontology engineering tasks. The thick arrow signs indicate dependency support and the thin arrow signs in the core ontology processing module point to sub-modules related to different functionalities.}
\label{fig:deeponto}
\end{center}
\end{figure}

Figure \ref{fig:deeponto} depicts the architecture of \deeponto. It consists of an ontology processing module as the foundation, which can support basic operations like loading, saving, retrieving entities,  querying for ancestors/descendants, and modifying entities and axioms, as well as more advanced functions like reasoning, verbalisation, normalisation, and projection (see Section \ref{sec:onto-proc} for their specifications). Built upon this basic module, \deeponto features a collection of ontology engineering tools and resources, devised for ontology alignment, completion, and ontology-based LM probing. \deeponto provides a fairly flexible and extensible interface for further implementations. This includes our ongoing efforts, such as logic embedding \cite{jackermeier2023box} and the identification and insertion of new concepts \cite{dong2023reveal}, as well as other typical works that can facilitate ontology construction and curation, such as the OWL ontology embedding method OWL2Vec* \cite{chen2021owl2vec}. The incorporation of these new tools and resources may necessitate the integration of other fundamental features into the core ontology processing module, further enhancing its capabilities and robustness. Through this positive development cycle, we expect \deeponto to emerge as a powerful package for the community, providing general support and fostering innovation within the field.

\section{Design Principle}

\subsection{Dependencies}

We chose the OWL API as the backend dependency due to its stability, reliability, and widespread adoption in notable projects and tools, such as Prot\'{e}g\'{e} \cite{musen2015protege}, ROBOT \cite{jackson2019robot} and HermiT \cite{glimm2014hermit}. \deeponto was initially built on Owlready2\footnote{\url{https://github.com/pwin/owlready2}} \cite{lamy2017owlready}, a Python-based ontology API. However, we found that Owlready2 is still in a preliminary stage and lacks support for several fundamental features. For instance, runtime errors are frequently triggered when attempting to delete entities from an ontology. Furthermore, it also posed difficulties when dealing with multiple ontologies simultaneously, as entities are loaded into a shared space without a reference to their original ontologies.
While the underlying library RDFLib\footnote{\url{https://rdflib.dev/}}, on which Owlready2 is built, does offer a platform to process RDF triples directly and introduce some missing functionalities for handling OWL ontologies, its use would necessitate significant additional development. This makes RDFLib a less viable option considering the developmental effort and resources required.
To facilitate the import of OWL API, we adopted the solution from the mOWL\footnote{mOWL is not used as a direct dependency but parts of its functionalities (e.g., normalisation and projection) encapsulated in Java were migrated to \deeponto to avoid duplicated implementations.} toolkit \cite{zhapa2023mowl}, which utilises JPype to connect the Java Virtual Machine (JVM) with Python programming. Specifically, the Java dependency files are compiled into the Java Archive (JAR) format and will be shipped upon package installation. JVM can then be launched within a Python program, establishing a connection to the JAR files.
We constrain the direct import of the Java dependencies within the central ontology processing module, maintaining them in a relatively static version. This strategy ensures long-term stability and streamlines the updating and management processes of the codebase.

\deeponto adopts PyTorch \cite{paszke2019pytorch} as the backbone for deep learning dependencies. PyTorch is characterised by its dynamic computation graph, which enables runtime modification of the model's architecture, providing flexibility and ease of use for users. Also, significant efforts are invested to ensure backward compatibility. Currently, ontology engineering modules in \deeponto mainly target applications of language models (LMs), which are well supported by the Huggingface's Transformers library \cite{wolf2019huggingface}. Building upon this, the OpenPrompt library \cite{ding2021openprompt} supports the prompt learning paradigm, which is one of the key foundations of recent cutting-edge large language models, such as ChatGPT~\cite{OpenAI2023GPT4TR} and LLaMA 2~\cite{touvron2023llama}.
\subsection{Architecture}

The architecture of \deeponto is straightforward and succinct. As shown in Figure \ref{fig:deeponto}, the basis of \deeponto is the core ontology processing module, which comprises a collection of essential sub-modules that revolve around the main class\footnote{In this work, the term \textit{``class''} refers to a blueprint for creating objects--encapsulating attributes and methods in Python programming.} \inlinecode{Ontology}.
The ontology class serves as the main entry point for introducing the OWL API's features, such as accessing ontology entities, querying for ancestor/descendent (and parent/child) concepts, deleting entities, modifying axioms, and retrieving annotations. These functions are encapsulated in a more cohesive and easy-to-use\footnote{For instance, when using the OWL API, retrieving subsumption axioms for different entity types requires distinct codes.  We have consolidated these similar functions into a single Python method for easier use.} manner. Along with these basic functionalities, we introduce several essential sub-modules to enhance the core module, such as reasoning, pruning, verbalisation, normalisation, projection, and more. One shared objective of all these components is to transform an ontology into various data forms, such as verbalising an ontology entity into natural language text and projecting an ontology into an RDF graph, thereby facilitating deep learning-based ontology engineering solutions. 

The core ontology processing module 
paves the way for implementing individual tools and resources to support ontology construction and curation. At present, \deeponto mainly incorporates systems based on pre-trained LMs, and has covered the tasks of ontology alignment, ontology completion with subsumptions, and ontology-based LM probing. In the near future, \deeponto is expected to incorporate new tools for ontology embedding, concept insertion, and more. Notably, the \deeponto system can be easily updated and extended with new tools to support other knowledge engineering scenarios, such as entity linking, entity alignment, and link prediction for knowledge graphs mainly composed of relational facts.

\section{Ontology Processing}\label{sec:onto-proc}

\begin{figure}[t]
    \centering
    \begin{minipage}{0.5\textwidth}
        \centering
        \includegraphics[width=0.98\linewidth]{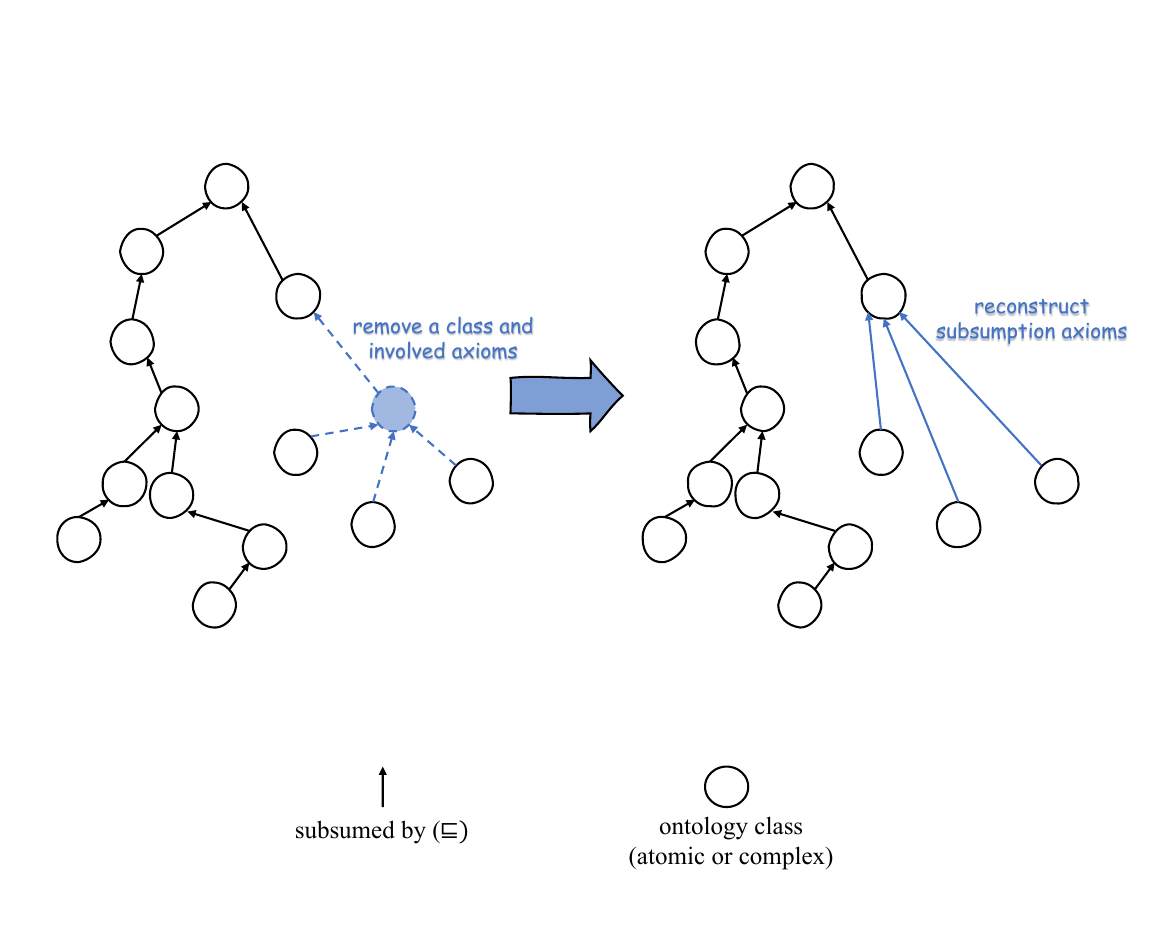}
    \end{minipage}%
    \begin{minipage}{0.5\textwidth}
        \centering
        \includegraphics[width=0.8\linewidth]{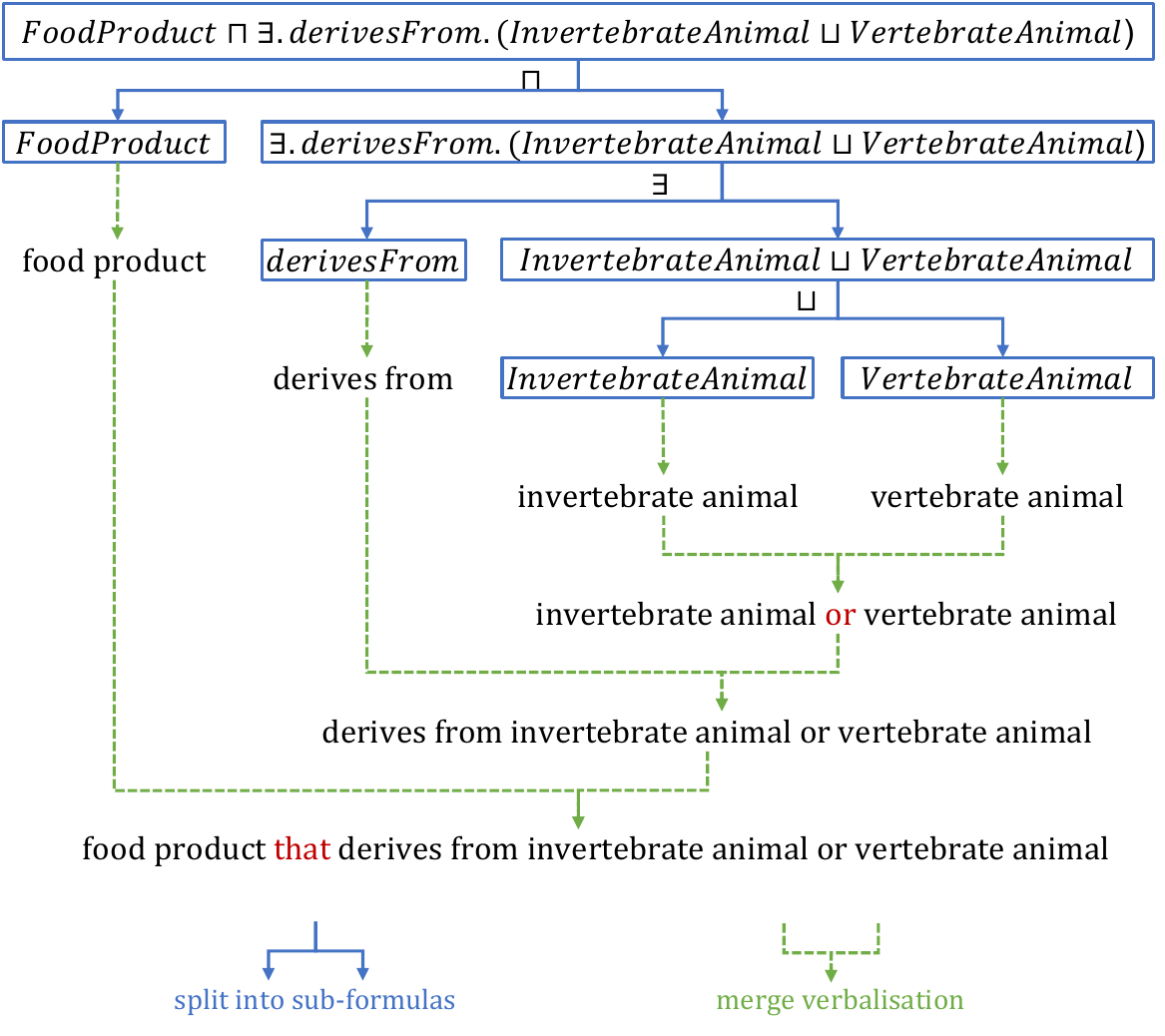}
    \end{minipage}
    \caption{The left figure illustrates the process of removing a concept while preserving the subsumption hierarchy in the ontology pruning algorithm proposed in Bio-ML~\cite{he2022bioml}. The right figure illustrates an example of the application of the recursive concept verbalisation algorithm proposed in OntoLAMA~\cite{he2023ontolama}.}
    \label{fig:pruning_and_verbalisation}
\end{figure}

In this section, we present a brief description for each component in the core ontology processing module. 

\subsubsection*{Ontology}
 
The base class of \deeponto is named as \inlinecode{Ontology}, which offers basic operations for accessing or modifying an ontology. An instance of \inlinecode{Ontology} can be initialised by taking an ontology file as input. Users can then access named entities (concepts and properties) through their IRIs, obtain asserted parents and children of an entity, retrieve and modify axioms, and so on. When implemented in Java using OWL API, even simple features like these may require several lines of code. \deeponto's encapsulation in Python improves code cleanliness and readability.

\subsubsection*{Ontology Reasoning}
Every instance of \inlinecode{Ontology} is accompanied by an instance of \inlinecode{OntologyReasoner} as its attribute. It is used for conducting reasoning activities, including obtaining inferred subsumers and subsumees, as well as checking entailment and consistency. By encapsulating these basic reasoning functions, we can implement more complex or specific reasoning algorithms. For example, we have implemented the \textit{assumed disjointness} proposed by \cite{he2023ontolama}, which can be particularly useful in the negative sampling process frequently employed in various machine learning-driven ontology curation tasks, including but not limited to, subsumption prediction. Currently, \deeponto supports three types of reasoners:

\begin{itemize}
    \item \textbf{HermiT} \cite{glimm2014hermit}, a sound and complete OWL reasoner based on hypertableau calculus. 
    \item \textbf{ELK} \cite{KKS12:elkimpl}, a highly optimised reasoner tailored to the OWL 2 EL profile.
    \item \textbf{Structural}\footnote{\url{https://owlcs.github.io/owlapi/apidocs_4/org/semanticweb/owlapi/reasoner/structural/StructuralReasoner.html}}, a simple structural reasoner.
\end{itemize}

\noindent Notice that \deeponto can be easily extended to incorporate other reasoners\footnote{E.g., \url{https://owlapi.sourceforge.net/reasoners.html}} that inherit the reasoner interface of OWL API.

\subsubsection*{Ontology Pruning} Real-world ontologies frequently exhibit a large-scale nature, leading to diminished efficiency during system evaluation. To extract a scalable subset from an ontology, we often prune the ontology by removing its concepts based on certain criteria, e.g., semantic types. In order to maintain the hierarchical structure during the pruning process, we implement the pruning algorithm proposed in \cite{he2022bioml} (found in \inlinecode{OntologyPruner}), 
which introduces subsumption axioms between the asserted (atomic or complex) parents and children of the concept targeted for removal (see Figure \ref{fig:pruning_and_verbalisation}; left). For future development, we aim to incorporate more pruning approaches. A key addition will be ontology modularisation, a technique that seeks to extract a (small) sub-ontology that entails a given axiom or is sufficient to answer a specific concept of queries~\cite{d2007ontology}. We also aim to explore its various variants that involve approximation, catering to different scenarios such as the construction of personalised knowledge graphs.

\subsubsection*{Ontology Verbalisation}  Verbalising an entity into natural language text with close meaning as its counterpart OWL statements can improve an ontology's accessibility and support many ontology engineering tasks. For example, ontology alignment systems often rely on string similarity or other text-level features to achieve successful results. While a named entity can be easily verbalised using its name (or labels), complex expressions that involve logical operators need a more sophisticated algorithm. In \deeponto, we implement a recursive concept verbaliser (found in \inlinecode{OntologyVerbaliser}) proposed in \cite{he2023ontolama}, which can automatically transform a complex logical expression into a textual sentence based on entity names or labels available in the ontology. An example is shown in the right of Figure \ref{fig:pruning_and_verbalisation}, where the complex concept expression $FoodProduct \sqcap \exists derivesFrom(InvertebrateAnimal \sqcup VertebrateAnimal)$ is parsed into a syntax tree of sub-formulas (this intermediate function is encapsulated in \inlinecode{OntologySyntaxParser}). The leaf nodes are named concepts or properties and they are verbalised directly. The recursion occurs when merging verbalised child nodes according to the logical pattern in their parent node, and terminates when the sentence is complete. In this example, the final output is \textit{``food product that derives from invertebrate animal or vertebrate animal''}. 

\subsubsection*{Ontology Normalisation}
Normalisation refers to the transformation of axioms into one of the following \textit{normal forms}: $C\sqsubseteq D$, $C\sqcap C'\sqsubseteq D$, $C\sqsubseteq\exists r. D$, $\exists r. C\sqsubseteq D$, $r \sqsubseteq s$, and $r \circ r' \sqsubseteq s$, with no loss of semantics~\cite{baader2005pushing}. Here, $C$ and $C'$ can be named concepts or $\top$, $D$ can be a named concept or $\bot$; and $r$, $r'$, and $s$ represent roles. The normalisation algorithm has been implemented in several $\mathcal{EL}$ embedding models \cite{kulmanov2019embeddings,jackermeier2023box} and this feature is also implemented in \deeponto for easy access. Normalisation can facilitate the training deep learning-based ontology engineering models because it simplifies the set of axiom patterns. 
For example, a complex subsumption $C\sqsubseteq D \sqcap \exists r. E$ can be normalised into $C \sqsubseteq D$ and $C \sqsubseteq \exists r. E$ such that a model trained on normalised axioms can make independent predictions for each axiom and then combine the results for a joint prediction.

\subsubsection*{Ontology Taxonomy} An ontology defines a hierarchy of concepts through asserted subsumption axioms, resulting in a taxonomy. This taxonomy simplifies the ontology’s structure, allowing easier navigation and retrieval by focusing exclusively on hierarchical relationships. This, in turn, facilitates graph-based deep learning models. In \deeponto, the ontology taxonomy (found in \inlinecode{OntologyTaxonomy}) is implemented as a directed acyclic graph with named concepts as nodes and subsumption relations as directed edges. The top concept \textit{owl:Thing} is used as the root node. To establish the subsumption edges, we employ an ontology reasoner (refer to the Ontology Reasoning section for more details) to infer \textit{direct} subsumers for each named concept. According to the OWL API documentation\footnote{\url{https://owlcs.github.io/owlapi/apidocs_5/org/semanticweb/owlapi/reasoner/OWLReasoner.html}}, a direct subsumption implies that if the ontology entails $C_1 \sqsubseteq C_2$ and there is no intermediate concept $C$ such that $C_1 \sqsubseteq C$ and $C \sqsubseteq C_2$, then $C_2$ is considered as a direct subsumer of $C_1$. Utilising a reasoner in this context helps to circumvent potential issues, such as mistakenly placing named concepts that are part of equivalence axioms directly under \textit{owl:Thing}. To illustrate, if $C_1 \equiv C_2 \sqcap C_3$ is the only axiom about $C_1$ asserted in an ontology, then the reasoner can infer $C_2$ and $C_3$ as direct subsumers of $C_1$, thus avoiding placing $C_1$ directly under \textit{owl:Thing} and making the taxonomy more complete.

\subsubsection*{Ontology Projection}
\deeponto offers the capability to transform an OWL ontology into a set of RDF triples. A default method for this transformation, which adheres to the W3C standard\footnote{\url{https://www.w3.org/TR/owl2-mapping-to-rdf/}}, is provided by OWL API. This method preserves the ontology's semantics and is effective for storing or exchanging ontologies. However, as it may introduce many blank nodes for representing complex logical expressions, its utility for ontology visualisation or applying graph-based algorithms, such as Random Walk and Graph Neural Networks, is limited. In such situations, a simplified graph representation is often needed, an approach sometimes termed as \textit{ontology projection}\footnote{Note that ontology taxonomy can be seen as a special case of projection, but we separate the functionalities into different modules because the former is purely for hierarchy and the latter is for triples.}. \deeponto has implemented the algorithm (found in \inlinecode{OntologyProjector}) originally used in the ontology visualisation system OptiqueVQS \cite{soylu2018optiquevqs}, to transform axioms into a set of simplified RDF triples.  
Briefly, a concept subsumption axiom $C \sqsubseteq D$ is transformed into $\langle C, rdfs\nsp:\nsp subClassOf, D \rangle$, an individual membership axiom $D(d)$ is transformed into $\langle d, rdf\nsp:\nsp type, D \rangle$, a role assertion axiom $r(a, b)$ is transformed into $\langle a, r, b \rangle$, a restriction axiom in the form of $C \sqsubseteq \exists r.D$ or $C \sqsubseteq \forall r.D$ is transformed into $\langle C, r, D \rangle$. Here, $C$ and $D$ are concepts; $a$, $b$, and $d$ denote individuals.

\section{Tools and Resources}

\deeponto has implemented several tools and resources for various ontology engineering purposes. For ontology matching (OM), \deeponto has BERTMap \cite{he2022bertmap} for concept equivalence matching, BERTSubs (Inter) \cite{chen2023contextual} for concept subsumption matching, and Bio-ML \cite{he2022bioml}, a collection of biomedical datasets and evaluation protocols to support OM system benchmarking. For ontology completion, \deeponto has BERTSubs (Intra), and the prompt-based approach proposed in OntoLAMA \cite{he2023ontolama}. The work of OntoLAMA also involves a collection of subsumption inference datasets for language model probing. In the following sub-sections, we present each of these modules in more detail.

\subsection{BERTMap}\label{sec:bertmap}

Ontology matching (OM) is 
the task of identifying mappings that represent a semantic relationship between entities of two different ontologies. BERTMap\footnote{BERTMap tutorial: \url{https://krr-oxford.github.io/DeepOnto/bertmap/}} targets on equivalence matching between named concepts. It adopts Bidirectional Encoder Representations from Transformers (BERT) \cite{devlin2018bert}, a masked language model pre-trained on extensive text corpora such as English Wikipedia, in the computation of a mapping score. Specifically, BERTMap utilises concept labels available in the input ontologies to extract pairs of synonyms and non-synonyms, and then fine-tunes a BERT model for synonym classification. The mappings score is determined by the aggregation of synonym scores between labels of two named concepts. To reduce the time complexity of mapping search, BERTMap adopts a sub-word inverted index for candidate selection. This approach takes advantage of the sub-word tokenisation capability inherent in the BERT model. To enrich the mapping set and capture potential missed matches during candidate selection, BERTMap incorporates an iterative mapping extension algorithm based on the locality principle, i.e., if two concepts are matched, their parents or children are likely to be matched. Lastly, BERTMap utilises the mapping repair module proposed in \cite{jimenez2011logmap} to remove a minimal set of inconsistent mappings. 

BERTMap provides flexible configurations to accommodate various needs. For instance, users can easily switch between different masked language models, such as RoBERTa \cite{liu2019roberta} and ALBERT \cite{lan2019albert}, or opt for BERT variants pre-trained on specialised corpora, such as BioBERT \cite{lee2020biobert} and ClinicalBERT \cite{huang2019clinicalbert}, by simply altering the input name of the language model. Moreover, BERTMap supports both unsupervised and semi-supervised settings; the former relies solely on the input ontologies, while the latter can leverage a small number of provided mappings for enhanced performance. Data augmentation from external ontologies is also possible and would be very helpful when the input ontologies are short of concept labels. 
Last but not the least, a light-weight version of BERTMap called BERTMapLt. This version does not necessitate the BERT fine-tuning nor the mapping refinement, rendering it considerably more efficient. Despite its simplified operations, BERTMapLt can attain promising results on certain datasets.


\subsection{BERTSubs}

BERTSubs\footnote{BERTSubs tutorial: \url{https://krr-oxford.github.io/DeepOnto/bertsubs/}} \cite{chen2023contextual} aims at predicting \textit{(i)} the missing concept subsumptions within an OWL ontology for completion, and \textit{(ii)} the subsumptions between concepts from two OWL ontologies for alignment. In our latest implementation in \deeponto (since v0.7.0), the super-concepts in both situations can be either named concepts or complex concepts such as existential restrictions.
Following the architecture of BERTMap, BERTSubs also fine-tunes a pre-trained BERT (or one of its variants) together with an attached binary classifier which outputs a score for an input candidate subsumption. 
It extracts the existing subsumptions within the given ontology or ontologies for constructing positive samples, and replaces the super-concepts of these subsumptions by a randomly selected named concept (or complex concept if the original super-concept is complex) for constructing negative samples.

Each concept in a subsumption is transformed into a text sentence, and the subsumption is transformed into a sentence pair as the model input.
For a complex concept, we directly call the verbalisation function implemented in the ontology processing module of \deeponto (see Section \ref{sec:onto-proc}).
For a named concept, we have implemented three approaches (see details in \cite{chen2023contextual}) to generate its (context-aware) text sentence as the model input: 
\begin{itemize}
\item \textbf{Isolated Class (IC)} which directly uses the given concept's name according to a pre-defined annotation property like \textit{rdfs:label}; 
\item \textbf{Path Context (PC)} which extracts a subsumption path from the ontology's concept hierarchy, starting from the given super-concept up to the root (or starting from the given sub-concept down to a leaf), and concatenates the names of the concepts in the path, separated by some special token;
\item \textbf{Breadth-first Context (BC)} which traverses a set of neighbouring subsumptions of the given concept from the ontology's concept hierarchy, starting from the given super-concept up to the root (or starting from the given sub-concept down to a leaf) via breadth-first search, and concatenates the names of the concepts of these subsumptions in a specific way, separated by a special token.
\end{itemize}

\subsection{Bio-ML}\label{sec:bio-ml}

\begin{table}[!t]
    \centering

    \begin{tabularx}{0.9 \textwidth}{X c Y Y c Y c Y}
    \toprule
    
         & \textbf{Ontology Pair} & \textbf{Category} & \textbf{\#Concepts~($\equiv$)}& \textbf{\#Refs~($\equiv$)} & \textbf{\#Concepts~($\sqsubseteq$)} & \textbf{\#Refs~($\sqsubseteq$)} \\\midrule
         
        \multirow{2}{5em}{Mondo} 
         & OMIM-ORDO & Disease & 9,642-8,838 & 3,721 & 9,642-8,735 & 103\\
         & NCIT-DOID & Disease & 6,835-8,848 & 4,686 & 6,835-5,113 & 3,339\\
         
         \midrule
         
        \multirow{3}{5em}{UMLS} 
         & SNOMED-FMA & Body & 24,182-64,726 & 7,256 & 24,182-59,567 & 5,506\\ 
         & SNOMED-NCIT & Pharm & 16,045-15,250 & 5,803 & 16,045-12,462 &  4,225\\ 
         & SNOMED-NCIT & Neoplas & 11,271-13,956 & 3,804 & 11,271-13,790 & 213 \\
         
    \bottomrule
    \end{tabularx}
    \caption{Data statistics for ontology pairs in Bio-ML, including the data sources, ontology names, categories (semantic types) of preserved concepts during ontology pruning, numbers of named concepts and reference mappings in the equivalence ($\equiv$) and the subsumption ($\sqsubseteq$) settings.
    }
    \label{tab:bio-ml-stats}
    
\end{table}

Benchmarking has been a critical challenge that limits OM academic research and system development.
Thus, we added Bio-ML into \deeponto. It currently includes five OM datasets derived from biomedical ontologies, accompanied by a comprehensive OM evaluation framework. The involved biomedical ontologies include SNOMED-CT~\cite{snomed}, FMA~\cite{fma}, NCIT~\cite{ncit}, OMIM~\cite{omim}, ORDO~\cite{ordo}, and DOID~\cite{doid}, based on the human-curated thesaurus and alignment in the integrated ontologies, UMLS~\cite{bodenreider2004unified} and Mondo~\cite{vasilevsky2020mondo}. It aims to address several limitations of the existing OM datasets and evaluation settings, including the following:

\begin{itemize}
\item \textbf{Sub-optimal ground truth mappings}: Many OM datasets have incomplete ground truth mappings, and conventional evaluation metrics Precision, Recall, and F-score become biased towards high-precision and low-recall systems, as they are more likely to match the available ground truth mappings.
\item \textbf{Limited to equivalence matching}: The majority of existing OM datasets focus solely on matching equivalent concepts.
\item \textbf{Lack of support for machine learning-based systems}: Current OM datasets do not accommodate data splitting (e.g., training, validation, and testing) and the conventional evaluation metrics, which are based on the final output mappings, are inefficient for developing and debugging machine learning-based OM models.
\end{itemize}

To address these limitations, Bio-ML incorporates the following strategies:

\begin{itemize}
\item \textbf{Human-curated mappings}: Bio-ML utilises ground truth mappings from UMLS and Mondo, which have been curated and validated by experts. 
\item \textbf{Expanded evaluation metrics}: In addition to conventional metrics, Bio-ML proposes the use of ranking metrics such as Hits@K and MRR to provide a more comprehensive assessment of OM systems and support more efficient evaluation of machine learning-based OM systems.
\item \textbf{Subsumption matching}: Bio-ML extends the scope of matching beyond equivalence, including subsumption relationships; the subsumption mappings are generated from the ground truth equivalence mappings.
\item \textbf{Data split settings}: Bio-ML formulates different data split configurations (unsupervised and semi-supervised) to better support the development and evaluation of machine learning-based OM systems.
\end{itemize}

Furthermore, given that biomedical ontologies are often large-scale and some OM systems cannot deal with such ontologies or will cost a very long time for computation, Bio-ML employs a pruning algorithm (see Ontology Pruning in Section \ref{sec:onto-proc}) subject to semantic types, which allows for the extraction of scalable sub-ontologies. For subsumption matching, target concepts that appear in a ground truth equivalence mapping used to construct a ground truth subsumption mapping is purposely deleted in order to enforce direct subsumption matching. The resulting datasets can be downloaded through Zenodo\footnote{Bio-ML dataset download: \url{https://zenodo.org/record/6946466}}, the detailed instructions and data statistics (also shown in Table \ref{tab:bio-ml-stats}) are provided in \deeponto\footnote{Bio-ML instructions: \url{https://krr-oxford.github.io/DeepOnto/bio-ml/}}. Moreover, Bio-ML has been introduced as a new track\footnote{Bio-ML Track of the OAEI: \url{https://www.cs.ox.ac.uk/isg/projects/ConCur/oaei/}} of the Ontology Alignment Evaluation Initiative (OAEI 2022), and the corresponding result report is available \cite{oaei-results}. We will discuss more about its usage in the OAEI in Section~\ref{sec:bio-ml-usage}.

\subsection{OntoLAMA}

The investigation of a language model's comprehension of knowledge and reasoning capability is a widely discussed topic, commonly referred to as ``LMs-as-KBs'' \cite{petroni-etal-2019-language}. While most existing works on 'LMs-as-KBs' focus on knowledge graphs composed of relational facts (sometimes known as RDF triple knowledge graphs), OntoLAMA explores OWL ontologies which represent conceptual knowledge with logical reasoning supported. OntoLAMA is essentially a set of subsumption inference (SI) datasets concerning both atomic and complex concept expressions. These datasets are extracted from ontologies of various domains and scales. The probing method proposed in OntoLAMA is based on prompt learning, which is a paradigm used to effectively extract knowledge from language models through prompts. In this method, concept expressions are first transformed into natural language text using the recursive ontology verbaliser (see Ontology Verbalisation in Section \ref{sec:onto-proc}). Then, the verbalised phrases are wrapped into a template along with the prompt text. The language model's task is to classify if two concept expressions have a subsumption relationship given this prompt-enhanced text format. Using this approach, a significant and consistent improvement in performance has been observed even with a small number of training and development samples (in a few-shot setting). This outcome highlights the potential of LM-based ontology engineering works in the future, without the need for excessive training resources. 

To summarise, OntoLAMA contributes the following to \deeponto: \textit{(i)} a set of LM probing datasets\footnote{OntoLAMA dataset download is available at Zenodo: \url{https://zenodo.org/record/7700458}, and Huggingface: \url{https://huggingface.co/datasets/krr-oxford/OntoLAMA}.} extracted from ontology subsumptions, \textit{(ii)} a verbalisation algorithm that can handle complex concept expressions, \textit{(iii)} a method to use LMs with prompts to predict subsumptions without fully supervised fine-tuning.

\section{Impact and Use}


\deeponto has been gaining attention from both the industry and academia. 
On the industrial front, notable adoptions of \deeponto include its utilisation by Samsung Research UK for Digital Health Coaching\footnote{\url{https://research.samsung.com/sruk}} and Madrid Digital\footnote{\url{https://www.comunidad.madrid/en/servicios/sede-electronica/madrid-digital}}, where it played a crucial role in their Proof of Concept process. Our user feedback indicates \deeponto's primary applications within the health and biomedical sectors. From the academia side, \deeponto contributes to the Bio-ML track of the Ontology Alignment Evaluation Initiative (OAEI), and our tools and resources have facilitated numerous research works, evident in existing scholarly citations. For example, BERTMap serves as a typical baseline in subsequent BERT-based ontology alignment studies, including but not limited to Truveta Mapper \cite{amir2023truveta}, LaKERMap \cite{wang2023contextualized}, as well as in large language model-based frameworks such as Retrieve-Rank \cite{wang2023exploring} and LLMap \cite{he2023exploring}.

In the following sub-sections, we elaborate on two use cases of \deeponto, i.e., Digital Health Coaching in Samsung Research UK and the Bio-ML track of the Ontology Alignment Evaluation Initiative (OAEI).

\subsection{Digital Health Coaching}

\begin{table}
    \centering
    \renewcommand{\arraystretch}{1.2} 
    \begin{tabularx}{0.8\textwidth}{X l Y Y Y Y Y Y}
    \toprule
    & & \multicolumn{3}{c}{\textbf{With out-of-scope concepts }} & \multicolumn{3}{c}{\textbf{Without out-of-scope concepts}} \\
    \cmidrule(lr){3-5} \cmidrule(lr){6-8}
    \textbf{System} & & \textbf{Precision} & \textbf{Recall} & \textbf{F-score} & \textbf{Precision} & \textbf{Recall} & \textbf{F-score} \\ \midrule
    \multicolumn{2}{l}{BERTMap} \\
    \multicolumn{2}{l}{\hspace{3mm} $\lambda = 0.995$} &  0.805 & 0.883 & 0.842 & 0.837 & 0.883 & 0.859 \\
    \multicolumn{2}{l}{\hspace{3mm} $\lambda = 0.999$} & 0.922 & 0.820 & \textbf{0.868} & 0.937 & 0.820 & \textbf{0.874} \\
    \multicolumn{2}{l}{Sub-string Match} & 0.965 & 0.489 & 0.649 & 0.965 & 0.489 & 0.649\\
    \bottomrule
    \end{tabularx}
    \caption{Human evaluation results derived from the unanimous consensus of three domain experts, where a mapping is validated only when all experts are in complete agreement, for aligning medical concepts from NHS conditions to DOID concepts. Given DOID's primary focus on diseases, non-disease concepts are categorised as \textbf{out-of-scope}.}
    \label{tab:sruk-eval}
\end{table}

Digital Health Coaching is commonly required in health management. A paramount priority is centred on the delivery of \textbf{context-sensitive, personalised, and explainable health recommendations}, aiming to aid end-users in understanding and ameliorating their health conditions. Consider, for instance, individuals afflicted with Gastro-oesophageal Reflux Disease who are utilising Amitriptyline to alleviate back pain. The responsibility lies with the Digital Health Coach to understand that the use of Amitriptyline requires the simultaneous administration of another medication to safeguard the stomach, such as Antacids or Duloxetine. Therefore, it should deliver a recommendation that includes an explanation, such as \textit{``Given that Amitriptyline can exacerbate reflux issues, it is recommended to discuss your stomach condition with your doctor as preventative measures might be necessary.''} 

In health scenarios where explainability and in-depth reasoning are prioritised, fully automatic Deep Learning algorithms prove inappropriate due to inherent constraints. These include a dependency on vast datasets for optimal performance, an inability to reason abstractly, and a nature that is largely inscrutable to human comprehension. A rising trend to alleviate these issues involves integrating a symbolic knowledge base (e.g., OWL ontology) that facilitates rigorous and complex reasoning into the process. One of the tasks involves automatically identifying entities from semi-structured and/or unstructured data, and subsequently mapping them to the knowledge base. The BERTMap model, as discussed in Section~\ref{sec:bertmap}, is utilised and evaluated in aligning health and/or medical concepts extracted from the NHS conditions\footnote{A list of medical concepts maintained by the \textbf{N}ational \textbf{H}ealth \textbf{S}ervice in the UK is available at: \url{https://www.nhs.uk/conditions/}.} to concepts in the DOID ontology\footnote{DOID version IRI: \url{http://purl.obolibrary.org/obo/doid/releases/2022-02-21/doid.owl}}.

The NHS conditions (at the time of investigation) consist of $984$ concepts that cover health conditions, symptoms, medical treatments/tools, possible causes, and related daily-life situations. 
Each concept is associated with a web-page of detailed information. As the NHS conditions are essentially a collection of web-pages, we conducted simple transformation over them into a \textit{``flat''} ontology, i.e., each term is modelled as an OWL ontology concept but there is no subsumption relationship among them. The concept names and aliases were manually collected from the web-pages to serve as concept labels. The DOID ontology has $10,924$ concepts for the sake of providing consistency, reusability, and sustainability of descriptions of human disease terms. As the NHS conditions are not all about diseases, some of the concepts are considered as \textit{``out-of-scope''} (i.e., non-diseases). We then considered task settings with and without the out-of-scope concepts. Since there are no ground truth mappings available, the alignment results were evaluated by human curators (see Table \ref{tab:sruk-eval}) because there are no existing ground truth mappings available. From the results we can observe that BERTMap attains a promising F-score and a consistently better performance than the Sub-string Match baseline model, which considers two concepts $C$ and $D$ as matched if a label of $C$ is a sub-string of a label of $D$, or vice versa. $\lambda$ denotes the threshold for filtering BERTMap's output mappings, i.e., only mappings with scores $\geq \lambda$ will be preserved. Sliding $\lambda$ is essentially a trade-off of Precision and Recall. BERTMap achieves an F-score of $0.868$ if considering out-of-scope concepts and an F-score of $0.874$ if not. This slight difference indicates that BERTMap wrongly matches some of the non-disease concepts to the disease concepts. For example, the drug concept \textit{Asprin} is wrongly mapped to the disease concept \textit{Aspirin-induced Respiratory Disease}. A potential reason for such mismatches is that BERTMap relies on sufficient ontology concept labels for high performance -- but only DOID provides multiple labels for each concept. Although the overall efficacy of BERTMap is satisfying according to the project coordinator, these mismatches underscore the need for further improvement by, e.g., utilising pertinent contexts to disambiguate the concepts.

\subsection{Ontology Alignment Evaluation Initiative}\label{sec:bio-ml-usage}

\begin{table}
\centering
\begin{tabularx}{0.95\textwidth}{c Y Y Y Y Y Y Y Y Y Y} \toprule
 & \multicolumn{5}{c}{\textbf{Unsupervised (90\% Test Mappings)}}  & \multicolumn{5}{c}{\bf Semi-supervised (70\% Test Mappings)} \\ \cmidrule(lr){2-6} \cmidrule(lr){7-11}
\textbf{System}	& \textbf{Precision}	& \textbf{Recall} & \textbf{F-score} & \textbf{MRR} & \textbf{Hits@1} & \textbf{Precision}	& \textbf{Recall} & \textbf{F-score} & \textbf{MRR} & \textbf{Hits@1}  \\ \midrule
LogMap	& 0.827	& 0.498	& 0.622	& 0.803	& 0.742 & 0.783 &0.547 &0.644 &0.821 &0.743  \\ 
LogMap-Lite & 0.935	& 0.259 & 0.405 & - & - & 0.932	& 0.519	& 0.667 & - & - \\  
AMD	& 0.664	& 0.565	& 0.611 & - & -& 0.792 & 0.528 & 0.633 & - & -  \\ 
BERTMap	& 0.730	& 0.572	& 0.641	& 0.873	& 0.817 & 0.575 & 0.784 & 0.664 & 0.965 &0.947  \\ 
BERTMapLt &	0.819 & 0.499 & 0.620 & 0.776 & 0.729 & 0.775&0.713&0.743&0.900&0.876  \\
Matcha	& 0.743 & 	0.508	& 0.604 & - & - & 0.704&0.564&0.626 & - &-  \\ 
Matcha-DL	& - & -& - & - & - & 0.956&0.615&0.748&0.654&0.640  \\ 
ATMatcher & 0.940 & 0.247 & 0.391 & - & - &0.835&0.286&0.426& - & - \\ 
LSMatch &	0.650	& 0.221 & 	0.329 & - & - & 0.877&0.238&0.374& - & - \\ \bottomrule
\end{tabularx}
\caption{Equivalence matching results for OMIM-ORDO (Disease) in the Bio-ML track of OAEI 2022.}
\label{tab:bioml1}

\vspace{.4cm}

\centering
\begin{tabularx}{0.95\textwidth}{c Y Y Y Y Y Y Y Y} \toprule
 & \multicolumn{4}{c}{\bf Unsupervised (90\% Test Mappings)}  & \multicolumn{4}{c}{\bf Semi-supervised (70\% Test Mappings)} \\ \cmidrule(lr){2-5} \cmidrule(lr){6-9}
System	& MRR	& Hits@1 & Hits@5 &Hits@10 &  MRR	& Hits@1 & Hits@5 &Hits@10  \\ \midrule
Word2Vec+RF	& 0.512	& 0.368	& 0.694	& 0.834	& 0.577	& 0.433	& 0.773	& 0.880 \\ 
OWL2Vec*+RF&	0.603	&0.461&	0.782&	0.860&0.666&0.547&0.827&0.880 \\
BERTSubs (IC)&	0.530&	0.333&0.786&	0.948&0.638&0.463&0.859&0.953 \\ \bottomrule
\end{tabularx}
\caption{Subsumption matching results for SNOMED-NCIT (Neoplas) in Bio-ML track of OAEI 2022.}
\label{tab:bioml2}
\end{table}

The Ontology Alignment Evaluation Initiative (OAEI) is an internationally coordinated initiative aiming at performing systematic assessments of ontology alignment techniques, also known as ontology matching (OM). As part of the OAEI 2022, we have introduced a new track, Bio-ML, replacing the previous LargeBio track. As detailed in Section~\ref{sec:bio-ml}, the objective of Bio-ML is to furnish datasets and a comprehensive framework for the evaluation of both traditional and machine learning-based OM systems. 

The OAEI 2022 version of Bio-ML comprises two ontology pairs derived from Mondo and three pairs from UMLS (see data statistics in Table \ref{tab:bio-ml-stats}). Every pair is linked with two types of matching: \textbf{equivalence} and \textbf{subsumption}. The equivalence matching incorporates both \textbf{global matching} (evaluated using traditional metrics like Precision, Recall, and F-score) and \textbf{local ranking} (assessed using ranking metrics such as MRR and Hits@K). The subsumption matching, due to its inherent incompleteness of ground truth mappings, only has the local ranking setting. Moreover, for each matching setup, there are two settings for data split: \textbf{unsupervised} and \textbf{semi-supervised}. The unsupervised setting does not provide any training mapping, while the semi-supervised setup provides a small partition designated for training and development. Note that \deeponto contributes to the construction of Bio-ML, the implementation of BERTMap and BERTMapLt systems, and the evaluation workaround.

The results for OMIM-ORDO equivalence matching and SNOMED-NCIT (Neoplas) subsumption matching are illustrated in Table \ref{tab:bioml1} and Table \ref{tab:bioml2}, respectively. Full results can be accessed on the Bio-ML (OAEI 2022) webpage\footnote{Bio-ML Track (OAEI 2022) Results: \url{https://www.cs.ox.ac.uk/isg/projects/ConCur/oaei/2022/index.html\#results}}. The findings from the Bio-ML track can be summarised as: \textit{(i)} ML-based systems generally outperform others; \textit{(ii)} no single system leads in all tasks, indicating that the effectiveness of systems varies based on the specific task; \textit{(iii)} traditional OM systems struggle with the subsumption matching task, primarily due to their dependency on lexical similarity, which is very useful in equivalence matching but not that much in subsumption matching.

\section{Conclusion and Future Work}

In this system paper, we introduce \deeponto, a Python library designed to facilitate ontology engineering with deep learning. The package provides a broad spectrum of ontology engineering capabilities, enabling users to (semi\nobreakdash-)\hspace{0pt}automatically and efficiently deal with ontologies and develop novel systems. The library relies on Python programming in synergy with deep learning methodologies, with a particular emphasis on pre-trained language models. 
For ease of use, \deeponto has encapsulated basic ontology processing functions from OWL API, and implemented several essential components such as reasoning, verbalisation, pruning, taxonomy, and projection. Based on these features, \deeponto has integrated a range of tools and resources for different ontology engineering tasks such as ontology alignment and completion, with comprehensive tutorials and detailed instructions\footnote{See the ``TUTORIALS'' section at \url{https://krr-oxford.github.io/DeepOnto/}.} provided. Furthermore, we show evidence of \deeponto's successful applications in both industry and academia, with promising results reported in different application contexts. 

For future development, we aim to incorporate more automated ontology engineering tasks. These include, but are not limited to, embedding concepts and properties with both formal semantics and literals considered, placing new concepts derived from text mentions into an ontology, generating descriptions for concepts based on their contexts in an ontology, and identifying, generating, and placing appropriate common parents for concepts clustered under certain criteria. These enhancements will lead to more ontology processing requirements, all of which will be seamlessly encapsulated into our API.

\section{Acknowledgements}

This work was supported by Samsung Research UK (SRUK), and the EPSRC projects OASIS (EP/S032347/1), UK FIRES (EP/S019111/1) and ConCur (EP/V050869/1).



\nocite{*}
\bibliographystyle{ios1}           
\bibliography{bibliography}        

%

\end{document}